\documentclass[11pt]{article}

\usepackage[final]{acl}

\usepackage{times}
\usepackage{latexsym}

\usepackage[T1]{fontenc}
\usepackage[utf8]{inputenc}
\usepackage{microtype}
\usepackage{inconsolata}

\usepackage{graphicx}

\usepackage{subcaption}
\usepackage{booktabs}
\usepackage{algorithm}
\usepackage{algorithmic}
\usepackage{tcolorbox}

\tcbset{
  vlmexample/.style={
    colback=gray!3,
    colframe=black,
    width=\linewidth,
    boxrule=0.8pt,
    arc=2pt,
    left=6pt,
    right=6pt,
    top=6pt,
    bottom=6pt,
  }
}

\newenvironment{vlmexample}[1]
{
\begin{tcolorbox}[vlmexample,title=#1]
}
{
\end{tcolorbox}
}

\usepackage{listings}

\lstdefinestyle{vlm}{
  basicstyle=\ttfamily\footnotesize,
  breaklines=true,
  breakatwhitespace=true,
  columns=fullflexible,
  keepspaces=true,
  showstringspaces=false,
  frame=single,
  xleftmargin=0pt,
  xrightmargin=0pt
}

\tcbuselibrary{listings,skins,breakable}

\newtcblisting{prompt}[1][]{
  listing engine=listings,
  enhanced, breakable,
  colback=gray!10, colframe=gray!70,
  boxrule=0.5pt, arc=1mm, outer arc=1mm,
  left=2mm, right=2mm, top=1mm, bottom=1mm,
  #1, listing only,
  listing options={
    basicstyle=\small,
    keywordstyle=\bfseries\color{blue!70!black},
    morekeywords={
      \<system\>,\</system\>,
      \<user\>,\</user\>,
      \<explanation\>,\</explanation\>,
      \<answer\>,\</answer\>,
      \{image\}
    },
    alsoletter={\<,\>,/},
    literate=
      {<system>}{{\color{blue!70!black}<system>}}1
      {</system>}{{\color{blue!70!black}</system>}}1
      {<user>}{{\color{blue!70!black}<user>}}1
      {</user>}{{\color{blue!70!black}</user>}}1
      {<explanation>}{{\color{blue!70!black}<explanation>}}1
      {</explanation>}{{\color{blue!70!black}</explanation>}}1
      {<answer>}{{\color{blue!70!black}<answer>}}1
      {</answer>}{{\color{blue!70!black}</answer>}}1
      {\{image\}}{{\color{red!70!black}\{image\}}}1,
    breaklines=true,
    breakindent=0pt,
    breakautoindent=false,
    columns=fullflexible,
  }
}

\usepackage{amsmath}
\usepackage{amssymb}
\usepackage{mathtools}
\usepackage{amsthm}

\title{Selective Agent Guidance via Entropy: \\ Learning Autonomous Policies from Imperfect VLM Teachers}

\author{
  \textbf{Giovanni Bonetta\textsuperscript{1}}\thanks{Equal contribution.},
  \textbf{Matteo Merler\textsuperscript{1}}\footnotemark[1],
  \textbf{Davide Zago\textsuperscript{2}},
  \\
  \textbf{Rossella Cancelliere\textsuperscript{2}},
  \textbf{Bernardo Magnini\textsuperscript{1}}
\\
  \textsuperscript{1}Natural Language Processing Unit, Fondazione Bruno Kessler, Trento, Italy
\\
  \textsuperscript{2}Department of Computer Science, University of Torino, Torino, Italy
\\
    \textbf{Correspondence:} \texttt{gbonetta@fbk.eu}, \texttt{mmerler@fbk.eu}
}

\begin{document}
\maketitle
\begin{abstract}
    Vision-Language Models (VLMs) provide useful priors for interactive decision-making, but using them directly as policies is expensive and brittle: they must be queried at every step, do not improve from environment interaction, and can repeat systematic errors. 
    We study how to learn a cheap autonomous policy from an online, expensive, and imperfect but informative VLM teacher. 
    We propose SAGE \footnote{Code available at: \url{https://github.com/giobin/SAGE}}(\textbf{S}elective \textbf{A}gent \textbf{G}uidance via \textbf{E}ntropy), a framework that queries a VLM only when the learner is uncertain, executes the suggested action during training, and distills guidance into a lightweight Reinforcement Learning (RL) policy. 
    Because VLM advice is not always reliable, SAGE can weight teacher-action distillation using environment-derived advantages rather than treating all suggestions as equally useful.
    Across sparse-reward visual reasoning and navigation tasks, SAGE learns policies that act without VLM guidance at evaluation time and improves over unguided RL in several environments, including settings where the learned policy exceeds its VLM teacher. 
    The results show that selective guidance is most beneficial when the VLM can help the agent discover high-reward trajectories, and less useful when unguided exploration already succeeds or teacher actions do not lead to informative experience.
    SAGE also reduces VLM usage by prompting the teacher only on a fraction of training steps and requiring no VLM calls at deployment. 
    Overall, our results suggest that VLMs don't need to be used as fixed policies to be useful; they can instead act as temporary, imperfect sources of guidance whose value is tested and internalized through interaction.
\end{abstract}

\begin{figure*}[t!]
    \centering
    \includegraphics[width=\linewidth]{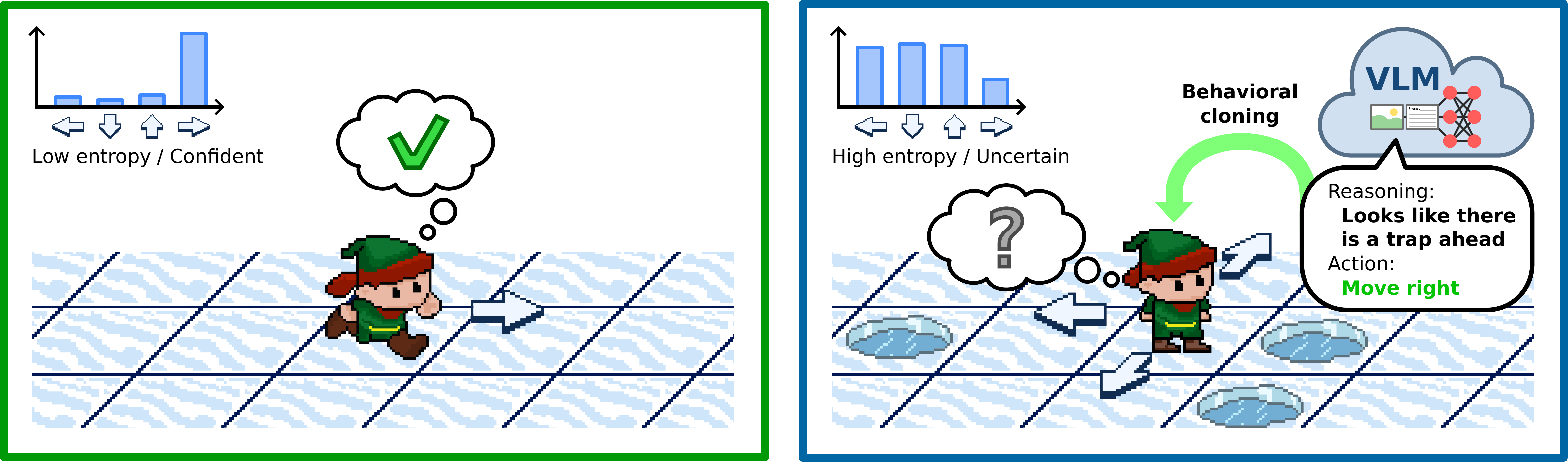}
    \caption{\textbf{Dual architecture for RL}. The agent is free to act in situations where it displays high confidence, but requests guidance from a VLM when it is uncertain, following the suggested action.}
    \label{fig:intro}
\end{figure*}

\section{Introduction}
Reinforcement Learning~\citep[RL;][]{sutton1998reinforcement} allows agents to improve through trial-and-error, but this comes with an important limitation: the agent typically starts tabula rasa, without prior knowledge of the environment. As a result, learning from sparse rewards can be extremely inefficient, since successful trajectories may be rarely discovered through random exploration~\citep{burda2018exploration}. This is particularly problematic in interactive environments that require perceptual grounding, symbolic reasoning, or multi-step planning: the agent may need to understand visual states and choose a sequence of correct actions before receiving any informative reward.

Vision-Language Models~\citep[VLMs;][]{liu2023visual, team2025gemma} offer a complementary set of capabilities. Through large-scale pre-training, they possess broad visual and linguistic priors that can help interpret objects, instructions, spatial relations, and symbolic structure. This makes VLMs attractive as interactive agents~\citep{ahn2022do, zhai2024finetuning}, but using them directly has drawbacks. A VLM must be prompted at every decision step, making deployment expensive and slow, and a frozen model does not improve through environment interaction; if it makes a systematic grounding or reasoning mistake, it may repeat that indefinitely.

This motivates a different problem setting. Rather than treating the VLM as the agent, we treat it as an online, expensive, and imperfect but informative teacher for a lightweight RL policy, used for targeted intervention during exploration. The VLM provides prior knowledge that can help the agent escape sparse-reward exploration bottlenecks, while RL provides the mechanism for testing suggested actions in the environment and improving beyond the teacher through trial-and-error. The goal is therefore to learn a cheap autonomous policy that uses the VLM during training, but acts without it at evaluation time. Since the teacher is not assumed to be correct, the learner should not rely only on blind imitation: teacher guidance should be combined with environment feedback so that useful interventions can be reinforced through RL.

We introduce SAGE (\textbf{S}elective \textbf{A}gent \textbf{G}uidance via \textbf{E}ntropy), a framework that uses a VLM as a temporary teacher for a lightweight RL agent. During training, the agent monitors the entropy of its own policy, which we use as a lightweight proxy for uncertainty. When this exceeds a threshold, the agent queries the VLM for an action and executes it in the environment. Teacher-guided actions are distilled into the learner, with an advantage-weighted objective that can emphasize actions associated with higher environment-derived returns.

We evaluate SAGE on sparse-reward visual decision-making environments that test complementary capabilities: FrozenLake and MiniGrid~\citep{chevalier2023minigrid} for visual navigation and object interaction, EZPoints~\citep{zhai2024finetuning} for arithmetic reasoning, CardMaze, a novel perceptual symbol-matching environment introduced in this work, and ALFWorld~\citep{shridhar2021alfworld} for grounded household interaction. Evaluation is performed without VLM guidance on held-out environment seeds, so performance measures what the lightweight policy has internalized rather than online VLM assistance.

Across these environments, SAGE can substantially improve over PPO where exploration bottlenecks dominate, including cases where the learned policy exceeds the VLM teacher itself. Compared to full-query methods such as VLM-as-policy, LVLM2P, and DAgger-style online imitation, SAGE queries the VLM on only a fraction of training steps and requires no VLM calls at deployment. Oracle-guided experiments further show that SAGE can exploit high-quality guidance to solve tasks that PPO fails to solve even after substantially longer training, while teacher-quality analyses clarify when failures arise from uninformative VLM guidance rather than the learning framework itself.

\section{Related Work}

\paragraph{Imitation learning and online supervision.}

Imitation learning has long been used to accelerate RL by transferring behavior from demonstrations to a learned policy~\citep{schaal1996learning, abbeel2004apprenticeship}. In Behavioral Cloning (BC), the learner matches expert state-action pairs, but can suffer from distribution shift when it visits states not covered by the demonstrations. Experts can take the form of human demonstrations~\citep{knox2009interactively}, pre-trained policies~\citep{rusu2015policy, parisotto16_actormimic}, or online teachers queried during learning. DAgger~\citep{ross2011reduction} addresses distribution shift by iteratively querying the expert on states visited by the learner and aggregating the resulting data, while interactive imitation learning more broadly studies settings where feedback is provided during policy execution~\citep{celemin2022interactive}. RCMP~\citep{dasilva2020uncertainty} uses uncertainty, measured by the standard deviation of multiple value heads, to query a stronger agent. SAGE differs from these methods as the online teacher is an expensive and imperfect VLM rather than a reliable expert, and we therefore query the teacher selectively and do not treat all guidance as equally trustworthy.

Because VLM teachers can be wrong, SAGE also relates to learning from imperfect demonstrations. AWR~\citep{peng2019awr}, AWAC~\citep{nair2020awac}, and CRR~\citep{wang2020critic} weight actions by estimated value, allowing suboptimal prior behavior to help learning without cloning every action equally. SAGE applies the same principle in an online VLM-guided RL setting: supervision is acquired on demand from the teacher in uncertain states, executed in the environment, and distilled according to environment-derived advantages.

\paragraph{Foundation models for interactive decision-making.}

Foundation models can be used in interactive environments either as sources of learning signals, as policies, or as teachers for smaller agents. One line of work uses Large Language or Vision-Language Models to provide feedback for RL, including preference labels~\citep{du2023guiding}, scalar rewards~\citep{kwon2023reward,ye2024reinforcement}, code reward functions~\citep{ma2023eureka, venuto2024code}, embedding-based rewards~\citep{klissarov2023motif, rocamonde2024visionlanguage}, or learned reward models~\citep{ma2022vip, ma2023liv}. RL-VLM-F~\citep{wang2024rlvlmf}, which we use as a baseline, prompts a VLM to compare pairs of transitions and trains a reward model from these preferences. SAGE instead uses the VLM to provide action-level guidance, allowing the teacher to directly help the learner reach sparse rewards while leaving the final policy to be optimized through RL.

Another line of work uses foundation models directly as policies~\citep{cao2024survey}, either by prompting them to output actions~\citep{ahn2022do, huang2022language, huang2022inner, ye2024reinforcement}, generating code policies~\citep{liang2023code}, or adding actions as a dedicated modality in Vision-Language-Action models~\citep[VLA;][]{brohan2023rt, kim2024openvla}. Foundation models can also be fine-tuned through environment interaction~\citep{Carta2023, tan2024true}, including VLMs~\citep{Bonetta24}; RL4VLM~\citep{zhai2024finetuning} trains VLM agents with PPO and chain-of-thought reasoning~\citep{wei2022chain, kojima2022large}. SAGE takes a different approach: the lightweight RL policy remains the deployed agent, while the VLM is used only as a temporary training-time teacher. This avoids querying or fine-tuning an expensive model at every control step while still leveraging its visual and linguistic priors.

Closest to our work are methods that distill foundation-model supervision into smaller policies. LM4TEACH~\citep{zhou2024large} uses an LLM expert and aligns the RL policy with token logits corresponding to actions, while LVLM2P~\citep{lee2025sample} asks a VLM to generate action probability distributions in text and distills them into a policy. SAGE differs by asking the VLM for a single action, which is a simpler generation task, querying the teacher only in high-entropy states, and training the final policy to act without the teacher at evaluation time. A preliminary study by \citet{merler2025guiding} explores uncertainty-triggered VLM guidance for RL, showing that VLM queries can decrease as policy entropy falls. SAGE studies whether such guidance can be converted into a generalizing autonomous policy: decreasing entropy alone is not sufficient, since policies can become confident without becoming competent. We therefore add explicit distillation from guided actions, validate performance on held-out tasks, and use oracle and ablation studies to isolate the role of teacher quality and policy learning.

\section{Method}
\label{sec:method}

\begin{figure*}[t]
    \centering
    \includegraphics[width=\linewidth]{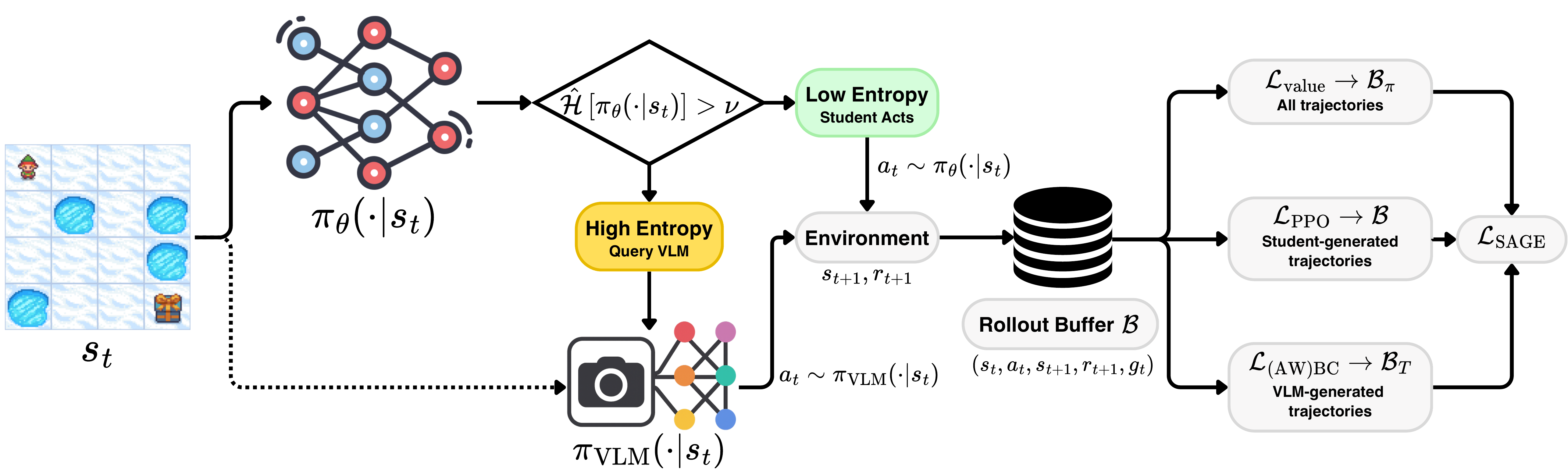}
    \caption{\textbf{Overview of SAGE.} At each training step, the student acts when its normalized policy entropy $\hat{\mathcal{H}}$ is below the threshold $\nu$; otherwise, SAGE queries the frozen VLM teacher and executes its suggested action. Transitions are marked with $g_t$ and partitioned into student-generated $\mathcal{B}_{\pi}$ and teacher-guided $\mathcal{B}_{T}$ subsets. PPO updates the actor on $\mathcal{B}_{\pi}$, BC uses teacher actions from $\mathcal{B}_{T}$, and the value function is trained on the full buffer $\mathcal{B}$.}
    \label{fig:pddl_diagram}
\end{figure*}

\subsection{Problem Setup}
\label{subsec:prelim}

We model the environment as a Markov Decision Process (MDP)
$\mathcal{M} = \langle \mathcal{S}, \mathcal{A}, R, T, \gamma, \rho_0 \rangle$,
where $\mathcal{S}$ is the state space, $\mathcal{A}$ is a discrete action space, $R$ is the environment reward, $T$ is the transition function, $\gamma$ is the discount factor, and $\rho_0$ is the initial state distribution. At each timestep, the learner observes a state $s_t$ and samples an action from a stochastic policy $\pi_\theta(a_t \mid s_t)$, trained with PPO~\citep{schulman2017proximal}. In our experiments, $s_t$ includes an RGB image and, when available, textual task information.

In addition to the agent, we assume access to an online teacher policy $\pi^T(\cdot \mid s_t)$ implemented by a VLM. The teacher can be queried in any state but has three limitations: it is expensive to call, it is not updated through environment interaction, and it may return incorrect actions, although some level of task-competence is needed in order to have some signal to follow. Our goal is to learn a cheap policy $\pi_\theta$ that maximizes environment return at evaluation time without querying $\pi^T$ (and while keeping the teacher frozen), while minimizing the number of teacher queries used during training.

\subsection{Entropy-Gated Selective Guidance}
\label{subsec:entropy_guidance}

We use the \textbf{entropy} of the agent's policy as a lightweight proxy for uncertainty:
\begin{equation}
\begin{aligned}
H_t
&= H\!\left[\pi_\theta(\cdot \mid s_t)\right] \\
&= - \sum_{a \in \mathcal{A}}
\pi_\theta(a \mid s_t)
\log \pi_\theta(a \mid s_t).
\end{aligned}
\end{equation}
which is normalized as~$\hat{H}_t = \frac{H_t}{\log |\mathcal{A}|}$, so that $\hat{H}_t \in [0,1]$ for any discrete action space. This is model-agnostic, requires no additional computation, and has been shown to provide a useful uncertainty signal in deep RL~\citep{sedlmeier2020policy}.

When $\hat{H}_t > \nu$, where $\nu$ is a fixed \textbf{threshold}, the learner queries the VLM teacher and executes the returned action. Otherwise, the learner samples from its own policy:

\begin{equation}
a_t \sim
\mu(\cdot \mid s_t) =
\begin{cases}
    \pi^T(\cdot \mid s_t) & \text{if } \hat{H}_t > \nu, \\
    \pi_\theta(\cdot \mid s_t) & \text{otherwise.}
\end{cases}
\label{eq:policyguide}
\end{equation}

Each transition is marked with an indicator $g_t \in \{0,1\}$ denoting whether the executed action was provided by the teacher. We also maintain a cache $C$ that stores state-action pairs returned by the model, to further reduce the cost of prompting a VLM in identical states.

\begin{figure*}[t]
    \centering
    \includegraphics[width=\linewidth]{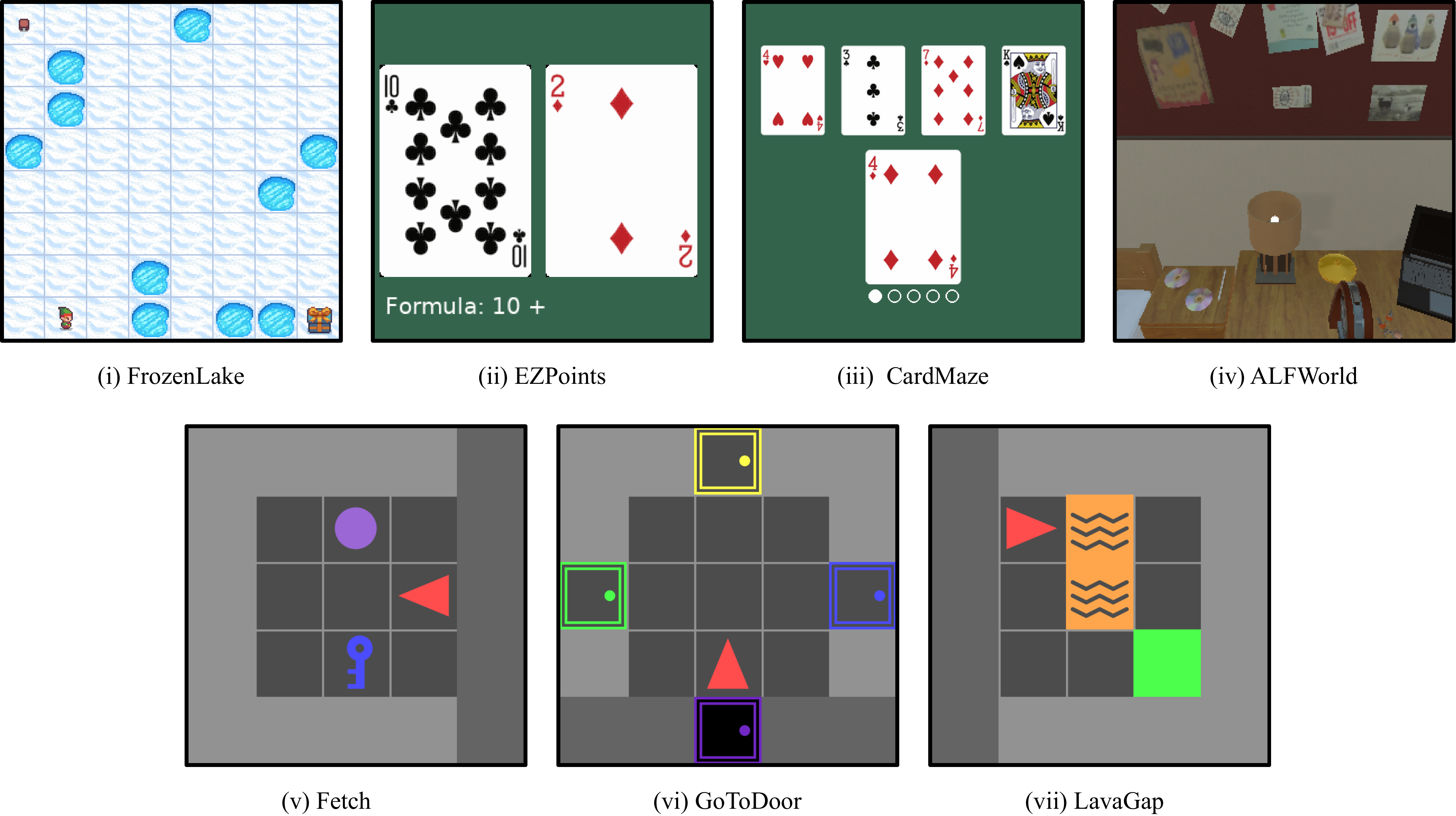}
    \caption{\textbf{RL environments used in our experiments.} (i) FrozenLake, (ii) EZPoints, (iii) CardMaze, (iv) ALFWorld, (v) MiniGrid Fetch, (vi) GoToDoor, and (vii) LavaGap.}
    \label{fig:environments}
\end{figure*}

\subsection{Partitioned Loss Functions}
\label{subsec:partitioned}

Teacher-guided actions are not sampled from the agent's policy, so treating them as ordinary PPO actions introduces an off-policy mismatch. This is especially problematic when the teacher selects actions that the learner assigns low probability: the resulting importance ratios can be large, and PPO clipping may overly suppress the signal.

We therefore partition the policy objective. Given the agent's current buffer of experience $\mathcal{B}$, let $\mathcal{B}_{\pi} = \{t \in \mathcal{B}: g_t = 0\}$ denote learner-generated transitions and $\mathcal{B}_{T} = \{t \in \mathcal{B}: g_t = 1\}$ denote teacher-guided transitions. Standard PPO policy updates (using the original clipped objective $\mathcal{L}_{\text{clip}}$) are applied only on $\mathcal{B}_{\pi}$:
\begin{equation}
    \mathcal{L}_{\text{PPO}}(\theta)
    =
    \mathbb{E}_{t \in \mathcal{B}_{\pi}}
    \left[
        \mathcal{L}_{\text{clip}}(\theta; s_t, a_t, \hat{A}_t)
    \right].
\end{equation}
Teacher-guided transitions are excluded from the PPO policy loss and instead contribute through the behavioral cloning objective described in Section~\ref{subsec:awbc}. This preserves the on-policy interpretation of PPO while allowing the teacher to influence the learner through a separate supervised signal.

We train the value function on both learner-generated and teacher-guided transitions. Although guided transitions affect the state distribution, the critic targets are computed from environment rewards rather than teacher labels. In sparse-reward tasks, this is important because teacher-guided actions may be the only way for the agent to observe successful trajectories early in training.

\subsection{Advantage-Weighted Behavioral Cloning}
\label{subsec:awbc}

A standard behavioral cloning loss would imitate all teacher actions:
\begin{equation}
    \mathcal{L}_{\text{BC}}(\theta)
    =
    -
    \mathbb{E}_{t \in \mathcal{B}_{T}}
    \left[
        \log \pi_\theta(a_t \mid s_t)
    \right].
\end{equation}
However, VLM teachers are imperfect. If the VLM suggests a visually plausible but incorrect action, blindly cloning this action can sharpen the learner policy around a mistake and reduce future uncertainty, making the agent less likely to ask for help in similar states while degrading the policy.

We therefore consider Advantage-Weighted Behavioral Cloning (AWBC), inspired by CRR~\citep{wang2020critic}:
\begin{equation}
    \mathcal{L}_{\text{AWBC}}(\theta)
    =
    -
    \mathbb{E}_{t \in \mathcal{B}_{T}}
    \left[
        w_t
        \log \pi_\theta(a_t \mid s_t)
    \right],
\end{equation}
where $w_t = f(\hat{A}_t)$ is a non-negative weight computed from the estimated advantage of the teacher-guided action. We use $f(\hat{A}_t) = \exp(\hat{A}_t / \tau)$ with temperature $\tau$, following CRR, and clip weights to $20$ for stability.

The advantage estimate is derived from the critic trained on environment rewards. Thus, AWBC uses environment feedback to modulate how strongly teacher actions are distilled: teacher actions that are estimated to lead to high-return trajectories receive larger weights, while actions associated with poor outcomes are down-weighted. Early in training, the critic may be uninformative, especially before any successful trajectory has been observed, so AWBC initially behaves similarly to unweighted BC rather than as a reliable filter of teacher quality. We therefore treat AWBC as an optional refinement to BC rather than as a guaranteed filter of teacher errors, and explore its effectiveness empirically.

The final SAGE objective is:
\begin{equation}
\begin{aligned}
\mathcal{L}_{\text{SAGE}}(\theta)
&=
\underbrace{
\mathcal{L}_{\text{PPO}}(\theta)
- c_H \mathcal{H}_{\pi}(\theta)
}_{\text{on } \mathcal{B}_{\pi}}
\\
&\quad
+ \underbrace{
\beta \mathcal{L}_{\text{AWBC}}(\theta)
}_{\text{on } \mathcal{B}_{T}}
+ \underbrace{
c_v \mathcal{L}_{\text{value}}(\theta)
}_{\text{on } \mathcal{B}} .
\end{aligned}
\label{eq:sage_objective}
\end{equation}
where $\mathcal{L}_{\text{value}}$ is the standard MSE loss with the environment reward, $\beta$ controls the strength of teacher distillation, $c_v$ is the value loss coefficient, and $c_H$ is the entropy regularization coefficient.

\section{Experimental Setup}
\label{sec:exp}

\subsection{Environments}
\label{subsec:environments}

Sample images representing a state $s_t$ for all environments are shown in Figure~\ref{fig:environments}.
\paragraph{FrozenLake 8$\times$8.}

We use a deterministic visual FrozenLake variant from Gymnasium~\citep{towers2024gymnasium}, with a new procedurally generated map each episode. The agent must reach the goal while avoiding holes and receives reward only upon success, preventing memorization of a fixed route.

\paragraph{MiniGrid.}

We evaluate on LavaGap, GoToDoor, and Fetch from MiniGrid~\citep{chevalier2023minigrid}. These tasks require visual navigation, obstacle avoidance when present, and object-centric interaction.

\paragraph{EZPoints.}
In EZPoints~\citep{zhai2024finetuning}, the agent must construct a formula that evaluates to $12$ using the values shown on two cards. The action space includes card values and arithmetic operators, and the agent must choose a valid sequence of actions to form the expression. Rewards are sparse, while illegal actions are penalized. 

\paragraph{CardMaze.}
We introduce CardMaze as a novel perceptual symbol-matching task. Each state contains four candidate cards and a prompt card. The correct action selects the candidate whose suit matches the prompt, with a confounder card sharing the same number. The agent receives reward only after $n=5$ consecutive correct selections. 

\paragraph{ALFWorld.}

ALFWorld~\citep{shridhar2021alfworld} is a grounded household interaction benchmark with visual observations, text goals, and admissible text actions. We report it as an exploratory stress test due to simulator and VLM inference cost.

\subsection{Baselines}
\label{subsec:baselines}

We evaluate SAGE against standard RL, direct VLM control, and representative VLM-assisted RL baselines. As a reference point, we consider a \textbf{PPO} agent trained directly on visual observations without any external guidance, measuring how much can be learned from sparse environment rewards alone. To characterize the standalone capabilities of the teacher, we also evaluate a \textbf{VLM-as-policy} baseline in which the VLM directly selects actions from image observations using CoT prompting. Unlike SAGE, this baseline keeps the VLM in the inference loop at every decision step and does not improve through environment interaction.

We then compare against methods that integrate VLMs into RL in different ways. \textbf{RL-VLM-F}~\citep{wang2024rlvlmf} uses the VLM for reward shaping by training a learned reward model from VLM-generated preferences over pairs of transitions. This tests whether VLM feedback is more effective as a dense reward signal rather than as action-level guidance. \textbf{LVLM2P}~\citep{lee2025sample} instead distills VLM predictions into the policy via supervised learning. In practice, this differs from SAGE in two important ways: the VLM is prompted at every timestep, and it is asked to generate a numerical probability distribution over actions rather than a single preferred action. We also implement a \textbf{DAgger-VLM} baseline~\citep{ross2011reduction} using the VLM as an online expert labeler: trajectories are collected under a mixture of the student and VLM policies, all visited states are labeled by the VLM, and the student is trained by BC on the aggregated dataset.

\subsection{SAGE Variants}
\label{subsec:our_variants}

To isolate the roles of guidance, distillation, teacher quality, and advantage weighting, we evaluate several variants of our method. Our full \textbf{SAGE} method uses entropy-gated VLM guidance, partitioned PPO updates, and AWBC.

We then consider \textbf{SAGE w/o BC}, which queries and executes the teacher in high-entropy states but removes the BC loss. This tests whether guided exploration alone is sufficient, or whether the policy needs an explicit supervised learning signal from teacher actions; similar to the work by~\citet{merler2025guiding}. We also evaluate \textbf{SAGE w/o AWBC}, which replaces AWBC with standard BC on guided transitions. This tests whether weighting teacher actions by environment-derived advantages is necessary, or whether standard distillation is sufficient once guidance is available.

Finally, we include \textbf{Oracle} variants that replace the VLM teacher with a rule-based oracle. This is not intended as a deployable baseline, but as a diagnostic to test how SAGE performs with a perfect teacher, to identify when performance is limited by teacher quality rather than by the learning framework itself.

\section{Results}
\label{sec:results}

We evaluate SAGE across four dimensions: overall performance, teacher quality, long-horizon convergence, and ablations of the learning objective.
All VLM-based methods use Qwen3.5-27B~\citep{qwen3.5} unless otherwise stated; ALFWorld uses Gemma3-27B~\citep{team2025gemma} and is reported as an exploratory evaluation. Learned policies are evaluated without VLM guidance on random seeds held out from training, so reported returns measure autonomous policy performance and generalization abilities rather than online VLM assistance. All environments are trained for 100k environment steps, except ALFWorld, which is trained for 40k steps due to simulator and VLM inference cost; furthermore, we find the ALFWorld oracle (as implemented by the original planner) to not always guarantee success. Because DAgger-VLM queries the VLM at every training step, we run it for 25k steps and report it as a cost-intensive online imitation baseline. We report prompts and hyperparameters in Appendices~\ref{app:prompts} and~\ref{app:hyperparam}, and a hyperparameter sweep in Appendix~\ref{app:hyperparam_sweep}.

\subsection{Main Results}
\label{subsec:main_results}

\begin{table*}[t]
\centering
\caption{\textbf{Evaluation performance comparison across environments.} Each cell shows the mean peak episodic return over 3 seeds with the standard deviation in parentheses. The best result per column is \textbf{bolded}; the best non-oracle result is \underline{underlined}.}
\label{tab:main_results}
\resizebox{\textwidth}{!}{
\begin{tabular}{lccccccc}
\toprule
\textbf{Method} & \textbf{FrozenLake} & \textbf{EZPoints} & \textbf{CardMaze} & \textbf{Fetch} & \textbf{GoToDoor} & \textbf{LavaGap} & \textbf{ALFWorld} \\
\midrule
PPO & $0.000$~{\scriptsize $(0.000)$} & $0.000$~{\scriptsize $(0.000)$} & $0.007$~{\scriptsize $(0.006)$} & $0.075$~{\scriptsize $(0.034)$} & $0.131$~{\scriptsize $(0.015)$} & \underline{$\mathbf{0.945}$}~{\scriptsize $(0.000)$} & $0.111$~{\scriptsize $(0.019)$} \\
VLM-as-Policy & $\underline{0.117}$~{\scriptsize $(0.065)$} & $\underline{0.175}$~{\scriptsize $(1.488)$} & $0.000$~{\scriptsize $(0.000)$} & $\underline{0.310}$~{\scriptsize $(0.044)$} & $0.127$~{\scriptsize $(0.019)$} & $0.083$~{\scriptsize $(0.143)$} & $0.108$~{\scriptsize $(0.000)$} \\
LVLM2P & $0.000$~{\scriptsize $(0.000)$} & $-2.868$~{\scriptsize $(2.618)$} & $0.000$~{\scriptsize $(0.000)$} & $0.036$~{\scriptsize $(0.040)$} & $0.020$~{\scriptsize $(0.026)$} & $0.000$~{\scriptsize $(0.000)$} & \textemdash{} \\
RL-VLM-F & $0.007$~{\scriptsize $(0.012)$} & $-0.450$~{\scriptsize $(0.507)$} & $0.037$~{\scriptsize $(0.064)$} & $0.062$~{\scriptsize $(0.006)$} & $0.111$~{\scriptsize $(0.040)$} & $0.190$~{\scriptsize $(0.329)$} & \textemdash{} \\
DAgger & $0.007$~{\scriptsize $(0.012)$} & $-3.970$~{\scriptsize $(0.165)$} & $0.993$~{\scriptsize $(0.012)$} & $0.099$~{\scriptsize $(0.000)$} & $0.052$~{\scriptsize $(0.028)$} & $0.000$~{\scriptsize $(0.000)$} & \textemdash{} \\

\textbf{SAGE (Ours)} & $0.103$~{\scriptsize $(0.111)$} & $0.000$~{\scriptsize $(0.000)$} & \underline{$\mathbf{1.000}$}~{\scriptsize $(0.000)$} & $0.122$~{\scriptsize $(0.025)$} & $\underline{0.147}$~{\scriptsize $(0.017)$} & $0.688$~{\scriptsize $(0.212)$} & \underline{$\mathbf{0.150}$}~{\scriptsize $(0.017)$} \\
\midrule
SAGE + Oracle & $\mathbf{0.697}$~{\scriptsize $(0.127)$} & $\mathbf{10.000}$~{\scriptsize $(0.000)$} & $0.663$~{\scriptsize $(0.574)$} & $\mathbf{0.456}$~{\scriptsize $(0.107)$} & $\mathbf{0.268}$~{\scriptsize $(0.029)$} & $\mathbf{0.945}$~{\scriptsize $(0.000)$} & $0.100$~{\scriptsize $(0.027)$} \\

\bottomrule
\end{tabular}
}
\end{table*}

Table~\ref{tab:main_results} reports the main performance comparison across all environments. SAGE improves over PPO in several environments, with the clearest gains on CardMaze, GoToDoor, Fetch, and the exploratory ALFWorld setting, while the remaining environments reveal limits caused by teacher quality (as discussed in Section~\ref{subsec:teacher_quality}) or already-strong baselines. Importantly, SAGE can also outperform the VLM teacher itself: on CardMaze, VLM-as-policy remains at $0.000$, while SAGE reaches the optimal return of $1.000$. This is the intended setting for SAGE: the VLM is not reliable enough to act directly as a policy, but its guidance is informative enough to help the RL agent discover sparse rewards and learn a stronger autonomous policy. DAgger also performs strongly on CardMaze, reaching $0.993$, which confirms that VLM supervision is useful in this environment; however, unlike SAGE, it relies on querying the VLM at every training step.

Instead, SAGE substantially reduces VLM usage, beyond the increases in performance. Across the six controlled environments, SAGE queries the VLM on only $1.2\%$--$13.3\%$ of training steps, concentrated in early training when entropy is still high, compared to $100\%$ for VLM-as-policy, LVLM2P, and DAgger. Query rates are lowest on FrozenLake, EZPoints, GoToDoor, Fetch, and LavaGap ($1.2\%$--$2.7\%$), and highest on CardMaze ($13.3\%$), where guidance remains useful for longer due to the larger state diversity. At deployment, SAGE requires no VLM calls, while direct VLM policies require one call at every decision step. Appendix~\ref{app:vlm_budget} reports the full per-environment query accounting and the cumulative VLM-query budget over training.

Among baselines, RL-VLM-F, LVLM2P, and DAgger struggle to achieve consistent gains across environments. RL-VLM-F depends on the VLM assigning informative preferences between visually similar transitions, which can be difficult in tasks where progress is sparse or visually subtle. LVLM2P is sensitive to generation format, since it requires the VLM to output a coherent numerical distribution over actions, whereas SAGE only asks for a single preferred action. DAgger avoids this distribution-generation issue, but still treats VLM labels as direct supervision and does not use environment rewards for policy optimization. In contrast, SAGE uses VLM actions as selective exploration interventions and lets environment feedback shape what is ultimately internalized.

The results also show that selective guidance is not uniformly beneficial. On LavaGap, PPO already solves the task, leaving little room for guidance to improve performance. On Fetch, the direct VLM policy is stronger than SAGE, suggesting that Qwen's immediate action choices are useful but are not yet converted into an equally strong learned policy under the current training setup. On FrozenLake, SAGE and VLM-as-policy obtain similar returns within seed variability, while PPO remains at zero; however, the low absolute performance suggests that Qwen guidance is still unreliable for precise grid localization. On EZPoints, both VLM-as-policy and SAGE perform poorly with Qwen3.5-27B, despite the task being solvable with oracle guidance. We analyze this behavior in Section~\ref{subsec:teacher_quality} as a teacher-quality failure case study.

Interestingly, SAGE + Oracle does not always outperform SAGE; we hypothesize that, while the oracle provides perfect signals, its deterministic nature may cause the agent to overfit to specific trajectories; conversely, the VLM's suboptimal stochasticity can act as a regularizer and avoids value overestimation in such cases.

\subsection{Effect of Teacher Quality} 
\label{subsec:teacher_quality} 

The main results show that SAGE's effectiveness depends not only on the learning algorithm, but also on the type of signal provided by the VLM teacher. To test whether direct-policy performance predicts usefulness as a teacher, we repeat the experiments with Gemma3-27B~\citep{team2025gemma} across all six controlled environments. Table~\ref{tab:teacher_quality} compares each VLM when used directly as a policy and when used to guide SAGE.

\begin{table*}[t]
\centering
\caption{\textbf{Effect of teacher quality across environments.} Results report peak episodic return as mean over 3 seeds, with standard deviation in parentheses. The best VLM-as-Policy result in each environment is \textit{italicized}, while the best SAGE result is \textbf{bolded}.}
\label{tab:teacher_quality}
\resizebox{\textwidth}{!}{
\begin{tabular}{lcccccc}
\toprule
\textbf{Usage} &
\textbf{FrozenLake} &
\textbf{EZPoints} &
\textbf{CardMaze} &
\textbf{Fetch} &
\textbf{GoToDoor} &
\textbf{LavaGap} \\
\midrule
\multicolumn{7}{l}{\textit{Qwen3.5-27B}} \\
VLM-as-Policy
& $\mathit{0.117}$~{\scriptsize $(0.065)$}
& $\mathit{0.175}$~{\scriptsize $(1.488)$}
& $0.000$~{\scriptsize $(0.000)$}
& $\mathit{0.310}$~{\scriptsize $(0.044)$}
& $\mathit{0.127}$~{\scriptsize $(0.019)$}
& $\mathit{0.083}$~{\scriptsize $(0.143)$} \\
SAGE
& $0.103$~{\scriptsize $(0.111)$}
& $0.000$~{\scriptsize $(0.000)$}
& $\mathbf{1.000}$~{\scriptsize $(0.000)$}
& $\mathbf{0.122}$~{\scriptsize $(0.025)$}
& $\mathbf{0.147}$~{\scriptsize $(0.017)$}
& $\mathbf{0.688}$~{\scriptsize $(0.212)$} \\
\midrule
\multicolumn{7}{l}{\textit{Gemma3-27B}} \\
VLM-as-Policy
& $0.000$~{\scriptsize $(0.010)$}
& $-3.400$~{\scriptsize $(0.070)$}
& $\mathit{0.670}$~{\scriptsize $(0.030)$}
& $0.000$~{\scriptsize $(0.000)$}
& $0.000$~{\scriptsize $(0.000)$}
& $0.000$~{\scriptsize $(0.000)$} \\
SAGE
& $\mathbf{0.127}$~{\scriptsize $(0.015)$}
& $\mathbf{10.000}$~{\scriptsize $(0.000)$}
& $\mathbf{1.000}$~{\scriptsize $(0.000)$}
& $0.059$~{\scriptsize $(0.010)$}
& $0.010$~{\scriptsize $(0.017)$}
& $0.000$~{\scriptsize $(0.000)$} \\
\midrule
\multicolumn{7}{l}{\textit{Random}} \\
SAGE
& $0.000$~{\scriptsize $(0.000)$}
& $0.000$~{\scriptsize $(0.000)$}
& $0.000$~{\scriptsize $(0.000)$}
& $0.059$~{\scriptsize $(0.000)$}
& $0.115$~{\scriptsize $(0.029)$}
& $0.000$~{\scriptsize $(0.000)$} \\
\bottomrule
\end{tabular}
}
\end{table*}

The comparison reveals that direct-policy and SAGE's performance are related, but not equivalent. EZPoints is the clearest example: Gemma gets a substantially lower return than Qwen when acting directly ($-3.400$ versus $0.175$), but SAGE reaches the optimal return of $10.000$ with Gemma, with Qwen-guided SAGE remaining at $0.000$, with a similar pattern shown by FrozenLake. On CardMaze, SAGE reaches optimal performance with either teacher despite a large difference in their direct-policy returns. Insead, on the three MiniGrid environments, Gemma is not effective both as a direct policy and as a teacher, while Qwen provides more useful guidance. Qualitatively, in FrozenLake Gemma often proposes sensible moves that bring the agent closer to the goal but does not complete the task due to its long-horizon, whereas in MiniGrid it frequently struggles with the action semantics and makes little progress toward the goal, suggesting uninformative actions instead. 

The random-teacher control further shows that a naive poor direct-policy does not result in useful guidance. Random actions never improve over PPO (Table~\ref{tab:main_results}) and can substantially degrade a policy that already learns from reward, as in LavaGap. Together, these results suggest that VLM performance alone is insufficient to predict teacher usefulness, but useful guidance must still contain task-relevant signal. SAGE therefore targets imperfect but informative teachers, rather than uninformative or systematically misleading ones.

\subsection{Ablation Study}
\label{subsec:ablation}

\begin{table*}[ht]
\centering
\caption{\textbf{Ablation study.}
Each cell reports peak episodic return as mean over 3 seeds, with standard deviation in parentheses.}
\label{tab:ablation}
\resizebox{\textwidth}{!}{
\begin{tabular}{lcccccc}
\toprule
\textbf{Method} & \textbf{FrozenLake} & \textbf{EZPoints} & \textbf{CardMaze} & \textbf{Fetch} & \textbf{GoToDoor} & \textbf{LavaGap} \\
\midrule
SAGE & $0.103$~{\scriptsize $(0.111)$} & $0.000$~{\scriptsize $(0.000)$} & $1.000$~{\scriptsize $(0.000)$} & $0.122$~{\scriptsize $(0.025)$} & $0.147$~{\scriptsize $(0.017)$} & $0.688$~{\scriptsize $(0.212)$} \\
SAGE w/o AWBC & $0.180$~{\scriptsize $(0.167)$} & $0.000$~{\scriptsize $(0.000)$} & $0.977$~{\scriptsize $(0.032)$} & $0.112$~{\scriptsize $(0.021)$} & $0.160$~{\scriptsize $(0.049)$} & $0.670$~{\scriptsize $(0.320)$} \\
SAGE w/o BC & $0.000$~{\scriptsize $(0.000)$} & $-2.410$~{\scriptsize $(3.408)$} & $0.000$~{\scriptsize $(0.000)$} & $0.060$~{\scriptsize $(0.010)$} & $0.069$~{\scriptsize $(0.036)$} & $0.000$~{\scriptsize $(0.000)$} \\
\midrule
SAGE + Oracle & ${0.697}$~{\scriptsize $(0.127)$} & ${10.000}$~{\scriptsize $(0.000)$} & $0.663$~{\scriptsize $(0.574)$} & ${0.456}$~{\scriptsize $(0.107)$} & ${0.268}$~{\scriptsize $(0.029)$} & ${0.945}$~{\scriptsize $(0.000)$} \\
SAGE + Oracle w/o BC & $0.000$~{\scriptsize $(0.000)$} & $-3.037$~{\scriptsize $(2.632)$} & $0.000$~{\scriptsize $(0.000)$} & $0.020$~{\scriptsize $(0.034)$} & $0.000$~{\scriptsize $(0.000)$} & $0.000$~{\scriptsize $(0.000)$} \\
\bottomrule
\end{tabular}
}
\end{table*}

We ablate the components that determine how teacher-guided actions are converted into policy learning: BC, advantage weighting, and teacher quality. Table~\ref{tab:ablation} shows that explicit BC is essential. Removing BC causes performance to collapse on all environments, even though teacher actions are still executed during training. The oracle rows strengthen this conclusion: even with a rule-based teacher, removing BC leads to near-zero performance across environments. Thus, guided exploration alone is not sufficient; teacher actions must also provide a direct policy-learning signal in order for them to be properly internalized.

In contrast, the comparison between SAGE and SAGE w/o AWBC shows no consistent benefit from advantage weighting: the two variants remain within seed variability across all six environments, with
plain BC outperforming AWBC in some cases. 
Thus, although AWBC provides a principled way to incorporate environment-validated information into teacher distillation, our experiments do not show that it is helpful. We thus view AWBC as an optional improvement; the essential component supported by the ablation is explicit BC distillation itself.

\subsection{Long-Horizon Evaluation}
\label{subsec:long_horizon}

The main results focus on the 100k-step setting, as VLM-guided interaction is costly. However, this does not fully distinguish if SAGE simply finds a solution faster (i.e., it is sample efficient), or if it converges to a better policy all together. To test this, we run a long-horizon experiment for 5M environment steps, using the oracle teacher for budget reasons, and to isolate whether the SAGE learning mechanism can exploit high-quality guidance when it is available.

\begin{table}[h]
\centering
\resizebox{\linewidth}{!}{
\begin{tabular}{lcc}
\toprule
\textbf{Environment} & \textbf{PPO} & \textbf{SAGE + Oracle} \\
\midrule
FrozenLake 
& $0.003$~{\scriptsize $(0.006)$} 
& $\mathbf{0.997}$~{\scriptsize $(0.006)$} \\
EZPoints 
& $0.000$~{\scriptsize $(0.000)$} 
& $\mathbf{10.000}$~{\scriptsize $(0.000)$} \\
CardMaze 
& $0.007$~{\scriptsize $(0.006)$} 
& $\mathbf{1.000}$~{\scriptsize $(0.000)$} \\
Fetch 
& $0.137$~{\scriptsize $(0.055)$} 
& $\mathbf{0.470}$~{\scriptsize $(0.208)$} \\
GoToDoor 
& $0.180$~{\scriptsize $(0.020)$} 
& $\mathbf{0.283}$~{\scriptsize $(0.116)$} \\
LavaGap 
& $\mathbf{0.945}$~{\scriptsize $(0.000)$} 
& $\mathbf{0.945}$~{\scriptsize $(0.000)$} \\
\bottomrule
\end{tabular}}
\caption{\textbf{Long-horizon evaluation over 5M environment steps.}
Results report peak episodic return as mean over 3 seeds, with standard deviation in parentheses.}
\label{tab:long_horizon}
\end{table}

Table~\ref{tab:long_horizon} shows that the benefit of guidance is not limited to early sample efficiency. On FrozenLake, EZPoints, and CardMaze, PPO remains near zero even after 5M steps, while SAGE + Oracle reaches near-optimal performance. This indicates that, in these sparse-reward environments, guidance can change which trajectories the agent discovers overall, not just how quickly it learns after reward has already been observed. On MiniGrid, PPO is stronger, especially on LavaGap, but oracle guidance still improves Fetch and GoToDoor.

\section{Conclusions}

We introduce SAGE, a framework for learning lightweight RL policies from an online, expensive, and imperfect but informative VLM teacher. Rather than using the VLM as the deployed policy, SAGE queries it selectively during training, executes teacher actions in high-entropy states, and distills useful guidance into the learner through a partitioned RL objective with BC, with the option to weight by advantage. This allows the VLM to act as a temporary exploration prior, while the final policy remains cheap and autonomous at evaluation time.

Across sparse-reward visual decision-making tasks, SAGE shows that selective VLM guidance can turn an imperfect teacher with some task-relevant competence into a useful training signal. It improves over unguided RL in several environments and can learn autonomous policies that outperform the VLM used to guide them. Our teacher-quality analysis shows that closed-loop VLM performance alone does not determine teacher usefulness: even weak direct policies can provide interventions that help agents discover high-reward behavior. These results support using VLMs not only as policies, but as temporary exploration guides for autonomous RL agents.

\section*{Limitations}

SAGE relies on policy entropy as a proxy for uncertainty when deciding whether to request guidance. This choice is simple and inexpensive, but it can conflate genuine uncertainty with action multimodality, and it does not directly estimate whether the teacher will be helpful in a given state. More expressive uncertainty estimates, such as ensembles, disagreement between value estimates, or learned query policies, may provide more precise control over when guidance is useful.

Our experiments focus on discrete action spaces. Extending SAGE to continuous control is conceptually possible, but on preliminary experiments we find current VLMs to perform poorly when asked to produce precise low-level numerical actions such as torques or velocities. A more realistic direction may be to use VLMs for high-level subgoals, skills, or action abstractions, or to replace the teacher with a Vision-Language-Action model trained for continuous control.

SAGE assumes that the teacher retains some task-relevant competence. As the random-teacher control and the LavaGap results show, uninformative or systematically misleading guidance can cause negative results, and SAGE is not designed to guarantee robustness in this regime, where we argue a teacher should not be used outright. Developing mechanisms that estimate teacher usefulness before executing or distilling an intervention is an important direction for future work.

Finally, while we include ALFWorld as a higher-dimensional household interaction stress test, this evaluation is preliminary and uses a smaller training budget than the controlled environments. Our controlled environments are designed to isolate the sparse-reward interactive exploration setting targeted by SAGE, rather than to model the full complexity of real-world embodied interaction. Larger-scale embodied experiments, stronger teachers, and more systematic comparisons to interactive imitation-learning baselines would further clarify where selective VLM guidance provides the largest advantage.

\paragraph{Potential Risks.}

SAGE may distill systematic biases or unsafe behaviors from the VLM teacher when flawed guidance leads to apparently successful trajectories. Although environment-derived advantage weighting can modulate the influence of teacher actions, our ablation does not show a consistent robustness benefit over plain BC, it does not guarantee that the learned policy is safe or aligned in settings where rewards are misspecified or incomplete, and it should always be tested by human evaluators before deployment, particularly in safety-critical applications.

\paragraph{Artifacts and Licenses.}

Our experiments build on publicly available environments and models, including Gymnasium, MiniGrid, ALFWorld, Qwen, and Gemma; their use is subject to the respective licenses and model terms. We will release code, prompts, and CardMaze assets under a permissive license, and will document third-party license requirements in the repository.

\section*{Acknowledgments}
Giovanni Bonetta and Bernardo Magnini were partially supported by the PNRR MUR project \href{https://fondazione-fair.it/}{PE0000013-FAIR} (Spoke 2).

\bibliography{custom}

\appendix
\clearpage

\section{Experiment Hyperparameters}
\label{app:hyperparam}

We report the hyperparameters used for SAGE and PPO training in our experiments on all
environments, in Tables~\ref{tab:SAGE_params} and~\ref{tab:ppo_params} respectively.

\begin{table*}[h]
\centering
\resizebox{\textwidth}{!}{
\begin{tabular}{lccccccc}
\toprule
\textbf{Parameter} & \textbf{FrozenLake} & \textbf{EZPoints} & \textbf{CardMaze} & \textbf{Fetch} & \textbf{GoToDoor} & \textbf{LavaGap} & \textbf{ALFWorld} \\
\midrule
Guidance VLM
& \multicolumn{6}{c}{\texttt{Qwen/Qwen3.5-27B}} & \texttt{google/gemma-3-27b-it} \\

Entropy threshold ($\nu$)
& 0.75 & 0.75 & 0.25 & 0.75 & 0.75 & 0.75 & 0.25 \\

BC coefficient ($\beta$)
& \multicolumn{7}{c}{1.0} \\

AWBC temperature ($\tau$)
& \multicolumn{6}{c}{0.5} & \textemdash{} \\

\bottomrule
\end{tabular}}
\caption{\textbf{SAGE hyperparameters across environments.} The six image/symbolic
environments use Qwen3.5-27B as the guidance VLM; ALFWorld preliminary experiments use Gemma-3-27B and no AWBC.}
\label{tab:SAGE_params}
\end{table*}

\begin{table*}[h]
\centering
\begin{tabular}{ll}
\toprule
\textbf{Parameter} & \textbf{Value} \\
\midrule
Agent type & CNN (+ LSTM text encoder for ALFWorld) \\
Convolution channels & \texttt{[16, 32, 32]} \\
Conv head out features & 256 \\
Actor std init & 0.01 \\
Critic std init & 1 \\
Gamma ($\gamma$) & 0.99 \\
GAE lambda ($\lambda$) & 0.95 \\
Value loss coef & 0.5 \\
Entropy coef & 0.01 \\
Clip coef & 0.2 \\
Clip value loss & \texttt{true} \\
Advantage normalization & \texttt{true} \\
Learning rate & $1\times10^{-3}$ \\
Learning rate annealing & \texttt{true} \\
Max gradient norm & 0.5 \\
PPO epochs & 8 \\
Minibatches & 4 \\
Batch size & 512 \\
\bottomrule
\end{tabular}
\caption{\textbf{PPO training hyperparameters.} ALFWorld uses a single environment
(\texttt{num\_envs}=1, batch size 64) with a CNN+LSTM text encoder; all other
environments use \texttt{num\_envs}=4 (batch size 512).}
\label{tab:ppo_params}
\end{table*}

\section{Hyperparameter Sweep}
\label{app:hyperparam_sweep}

We report a sweep of the AWBC weight ($\beta$) and the guidance threshold ($\nu$) on the CardMaze environment, with 3 seeds per value. Table~\ref{tab:sweep_beta} sweeps $\beta$ with $\nu = 0.25$ fixed, while Table~\ref{tab:sweep_nu} sweeps $\nu$ with $\beta = 1$ fixed. The results suggest that $\beta$ is best around $1$, while $\nu$ works well in a range of roughly $0.05$--$0.45$.

\begin{table}[h]
\centering
\small
\begin{tabular}{lcccc}
\toprule
$\beta$ & $0.0$ & $0.5$ & $1.0$ & $2.0$ \\
\midrule
Mean & $0.000$ & $0.663$ & $\mathbf{1.000}$ & $0.800$ \\
$\pm$ Std. & $0.000$ & $0.575$ & $0.000$ & $0.346$ \\
\bottomrule
\end{tabular}
\caption{\textbf{Sweep of the BC coefficient $\beta$ on CardMaze ($\nu = 0.25$).} Mean and standard deviation of final return over 3 seeds; best value in bold.}
\label{tab:sweep_beta}
\end{table}

\begin{table}[h]
\centering
\small
\setlength{\tabcolsep}{3.5pt}
\resizebox{\linewidth}{!}{
\begin{tabular}{lccccccc}
\toprule
$\nu$ & $0.00$ & $0.10$ & $0.25$ & $0.50$ & $0.75$ & $0.90$ & $1.00$ \\
\midrule
Mean & $0.503$ & $\mathbf{0.997}$ & $0.993$ & $0.477$ & $0.030$ & $0.090$ & $0.000$ \\
$\pm$ Std. & $0.500$ & $0.006$ & $0.012$ & $0.454$ & $0.044$ & $0.027$ & $0.000$ \\
\bottomrule
\end{tabular}}
\caption{\textbf{Sweep of the guidance threshold $\nu$ on CardMaze ($\beta = 1$).} Mean and standard deviation of final return over 3 seeds; best value in bold.}
\label{tab:sweep_nu}
\end{table}

\section{VLM Query Budget over Training}
\label{app:vlm_budget}

This appendix quantifies the training-time cost of SAGE in terms of VLM queries, complementing the query-rate discussion in Section~\ref{subsec:main_results}. During training we log every prompt actually issued to the VLM teacher; states served by the guidance cache (Section~\ref{subsec:entropy_guidance}) do not trigger new calls. Table~\ref{tab:vlm_budget} reports, for each environment, the total number of VLM calls accumulated over a full training run (mean and standard deviation over 3 seeds) and the corresponding query rate, i.e., the fraction of environment steps on which the teacher was prompted.

\begin{table}[t]
\centering
\small
\begin{tabular}{lrrr}
\toprule
\textbf{Environment} & \textbf{Steps} & \textbf{VLM calls} & \textbf{Rate} \\
\midrule
FrozenLake & 100k & $1{,}708 \pm 53$    & $1.7\%$ \\
EZPoints   & 100k & $1{,}160 \pm 278$   & $1.2\%$ \\
CardMaze   & 100k & $13{,}251 \pm 1{,}295$ & $13.3\%$ \\
Fetch      & 100k & $2{,}718 \pm 1{,}059$  & $2.7\%$ \\
GoToDoor   & 100k & $1{,}462 \pm 544$   & $1.5\%$ \\
LavaGap    & 100k & $2{,}042 \pm 309$   & $2.0\%$ \\
ALFWorld   & $\sim$40k & $10{,}118 \pm 260$  & $26.9\%$ \\
\bottomrule
\end{tabular}
\caption{\textbf{Training-time VLM query accounting for SAGE (mean $\pm$ standard deviation over 3 seeds).} The query rate is computed per run as the number of VLM calls divided by the environment steps of that run, then averaged over seeds. Full-query methods (VLM-as-policy, LVLM2P, DAgger) prompt the VLM at every training step.}
\label{tab:vlm_budget}
\end{table}

\begin{figure*}[t]
    \centering
    \includegraphics[width=\textwidth]{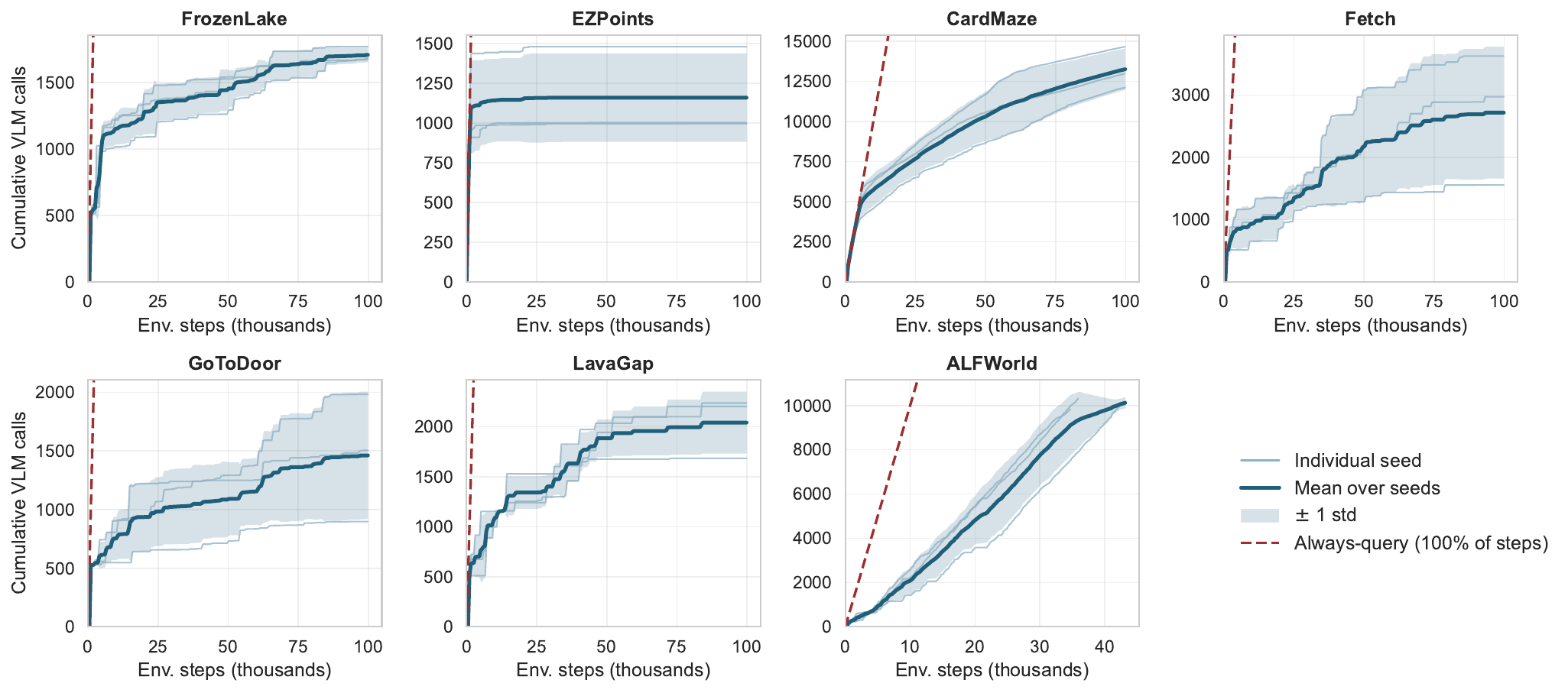}
    \caption{\textbf{Cumulative VLM-query budget over training.} Each panel shows the cumulative number of VLM prompt calls issued by SAGE as a function of environment steps, for individual seeds and their mean $\pm$ one standard deviation. The dashed red line is the budget of an always-query approach (one call per environment step), for reference. Because queries are triggered only in high-entropy states, queries are concenrated at the beginning, and tend to reduce as the policy becomes confident.}
    \label{fig:vlm_budget_curves}
\end{figure*}

Figure~\ref{fig:vlm_budget_curves} shows \emph{when} this budget is spent. In the six controlled environments, calls are concentrated in the earliest phase of training, when the policy entropy is still high: more than half of the total budget is typically spent well before the midpoint of training, and the curves progressively flatten as the entropy gate closes, with EZPoints plateauing almost immediately. CardMaze is the main exception: its larger state diversity keeps the policy uncertain on novel configurations for longer, so guidance continues to be requested throughout training. On ALFWorld, the query budget grows almost linearly, matching the higher query rate of this exploratory setting.

For comparison, full-query methods spend one call per step: at the same 100k-step budget, VLM-as-policy and LVLM2P issue $100{,}000$ calls per run, i.e., between $7.5\times$ (CardMaze) and $86\times$ (EZPoints) more than SAGE, and DAgger issues $25{,}000$ calls in its shortened 25k-step runs, still well above SAGE on every controlled environment. Moreover, these budgets do not include deployment: VLM-as-policy continues to pay one call per action after training, while the policies learned by SAGE act autonomously and issue no VLM calls at evaluation time.

\section{Qualitative Analysis and Failure Cases of VLM Guidance}
\label{app:fails}
We present some example failure cases for SAGE and LVLM2P, involving the VLM failing to generate the optimal action.

Examples A and B show sample responses from SAGE, with the VLM failing to properly understand the grid-world and assess spatial relationships. This could be due to the way the image is processed, or because pixel-art grid-worlds are not too prevalent in the model's pre-training distribution, making it unable to properly reason over exact coordinates.

Example C instead shows a sample response from LVLM2P-style prompting on EZPoints, highlighting the difficulty for VLMs to produce exact state distributions, particularly as the number of actions grows. SAGE mitigates this by only asking for the optimal action, rather than a full distribution.

\section{Computational Resources}

All experiments were ran on a cluster equipped with NVIDIA A100 GPUs with 64GB of VRAM. Experiments used two such GPUs when running with Qwen3.5-27B or Gemma3-27B, and one otherwise. Between preliminary experiments, and all runs reported in the paper, we estimate around 4000 GPU hours were used for our experiments.

\section{VLM Guidance Prompts}
\label{app:prompts}

We present the few-shot prompts with CoT used to query the VLM to guide the SAGE agent. The \{image\}, \{mission\}, \{direction\}, and \{admissible\_commands\} fields are replaced at runtime with the state $s_t$ and corresponding information.

\onecolumn
\begin{prompt}[title=Frozenlake]
<user>
{image}

You are a smart agent acting in the "FrozenLake" environment from the Gymnasium library. You will receive observations from the environment and must decide which action to take next.
The goal is to navigate across a frozen grid to reach the goal without falling into holes.

## Observation space
You are presented with the image of the environment in the style of pixel art.
Each cell may be safe (white with snow), a hole (bright blue ice hole), the start (a stool) or the goal (a present box). The agent is represented by an elf character.

## Action space
The action space includes 4 discrete actions:
- 0: Move left
- 1: Move down
- 2: Move right
- 3: Move up

# Output Format
You should think about the answer step by step inside the <think> </think> tags, then provide the chosen action index inside the <action> </action> tags. All text outside the <think> and <action> tags will be ignored.

--------------------------------
# Few-shot Examples (2-shot)
--------------------------------

## Example 1
## Observation:
image: [the image shows the elf in row 4, column 7 (i.e. the coordinates are from 1 to 8). There is a hole left of him and one below.]

<think>
The agent is currently at the center right corner of the grid. The goal is below the agent. The agent needs to avoid the hole and then proceed downwards toward the goal which is located at the bottom right corner. The possible moves to avoid the holes are "move up" and "move right". Moving up is safe but would result in the agent getting further away from the gift. Given all these considerations probably a good option is going up and reevaluate the situation, and a even better option is to go right.
Move right is action 2.
</think>
<action>2</action>

## Example 2
## Observation:
image: [the image shows the elf in row 3, column 3. There is a hole directly to the right, safe tiles straight down toward the bottom row, and the goal is further down to the right.]

<think>
The elf is a couple of tiles above the center of the grid. The cell to the right is a hole, so moving right (action 2) is risky. Moving left would take the agent back toward the start and does not progress toward the goal. Moving up only keeps the elf wandering on the same plateau. The only safe and progress-making move is to step down along the clear column toward the goal. Therefore, action 1 (move down) is preferred.
</think>
<action>1</action>
--------------------------------

# Current Task
## History:

Based on the current visual information, which action should the agent take next?

</user>
\end{prompt}

\begin{prompt}[title=EZPoints]
<user>
{image}

You are playing a card game. The goal is to output a math formula that evaluates to 12 using the numbers in the poker cards. The goal is to use all numbers in the cards (only once) to compute 12. At each state, only numbers that appear in the cards and math operators are legal actions. "J", "Q", or "K" are treated as "10", and ace cards are treated as "1". The game ends when the "=" operator is added to the formula.

# Workflow Checklist
- Begin each turn with a concise checklist: (1) analyze your current hand and the displayed formula, (2) identify all valid actions, (3) select the optimal move that advances towards a valid formula evaluating to 12, (4) output the corresponding action number.

## Observation space
You will receive an image displaying:
- The two cards currently in hand.
- The current, possibly incomplete formula (displayed after 'Formula:').

## Action space
The action space includes 13 discrete actions:
- 1: add number "1" to the formula
- 2: add number "2" to the formula
- 3: add number "3" to the formula
- 4: add number "4" to the formula
- 5: add number "5" to the formula
- 6: add number "6" to the formula
- 7: add number "7" to the formula
- 8: add number "8" to the formula
- 9: add number "9" to the formula
- 10: add number "10" to the formula
- 11: Add the operator "+" (addition) to the formula
- 12: Add the operator "*" (multiplication) to the formula
- 13: Add the operator "=" (equals) to the formula

# Output Format
You should think about the answer step by step inside the <think> </think> tags, then provide the chosen action index within the <action> </action> tags. All text outside the <think> and <action> tags will be ignored.

--------------------------------
# Few-shot Examples
--------------------------------

## Example 1
## Observation:
image: [the image shows two cards in hand: "J of clubs" and "2 of spades"; Formula: "" (empty)]

<think>
The cards in hand are a 2 of spades and a Jack of clubs. Since Jack is treated as 10, the numbers available are 2 and 10. The goal is to create a formula that evaluates to 12. 

Possible actions:
1. Add the number 2 to the formula.
2. Add the number 10 to the formula.
3. Add the addition operator (+).
4. Add the multiplication operator (*).
5. Add the equals operator (=) if a valid formula is formed.

Starting with the numbers:
- 2 and 10 can be combined with addition: 2 + 10 = 12.

The simplest formula is 2 + 10 = 12. Also 10 + 2 = 12 is valid.
Let's go with the first option, so add the number 2 to the formula.
Number 2 corresponds to action 2.
</think>
<action>2</action>

## Example 2
## Observation:
image: [the image shows two cards in hand: "7 of diamonds" and "5 of diamonds"; Formula: "10"]

<think>
Okay, let's see. The cards I have are a 7 of diamonds and a 5 of diamonds. The current formula is "7". I need to use both numbers to make the formula equal 12. 

So far, I have "7". If I add 5 next, that would make it "7 + 5", which equals 12. That works! I should add the 5 to the formula. 
To output add "5" to the formula i need to output the "+" symbol first. The "+" symbol corresponds to action 11.
</think>
<action>11</action>

## Example 3
## Observation:
image: [the image shows two cards in hand: "Q of hearts" and "2 of clubs"; Formula: "10"]

<think>
The cards in hand are the Queen of hearts and the 2 of clubs. The Queen is treated as 10. The current formula is "10". 

Possible actions:
1. Add the number 10 to the formula.
2. Add the number 2 to the formula.
3. Add the addition operator (+).
4. Add the multiplication operator (*).
5. Add the equals operator (=).

Since the current formula is "10", adding the number 2 will give "10 + 2". 

10 + 2 = 12, which is the target value.

So, since the formula is just "10", the next action should be to add the operator "+" to the formula.
The + operator corresponds to action index 11.
</think>
<action>11</action>
--------------------------------

# Current Task
## History:

Based on the current visual information, which action should the agent take next?
</user>
\end{prompt}

\begin{prompt}[title=CardMaze]
<user>
{image}

You are an expert agent playing a simple game of cards.
You will be presented with {n_cards} different poker cards at the top of the image. At the bottom of the image, you will be shown a larger poker card which acts as your prompt.
You will be given a task, which determines which card you should select from the top cards, based on the card shown in the prompt.
If you make a mistake, the progress will reset to zero and you will have to start over. The game ends when you have made {n_decisions} correct actions in a row.

# Workflow Checklist
- Begin each turn with a concise checklist: (1) analyze the image to identify which card is in the prompt and the top cards, (2) find which card at the top matches the prompt according to the task, (3) identify the index of the matching card, (4) select the action which corresponds to the index of the matching card.

# Output Format
You should think about the answer step by step inside the <think> </think> tags, provide the chosen action index inside the <action> </action> tags. All text outside the <think> and <action> tags will be ignored.

--------------------------------
# Few-shot Examples
--------------------------------

## Example 1
## Task
Your task is to produce an action corresponding to the card at the top that matches the suit (hearts, diamonds, clubs, spades) of the prompt card, but not its number or face.

## Action space
The action space includes 4 discrete actions:
- 0: Choose the 1st card from the left
- 1: Choose the 2nd card from the left
- 2: Choose the 3rd card from the left
- 3: Choose the 4th card from the left

## Observation:
image: [the image shows 4 cards in hand (top of the image): "J of clubs", "2 of spades", "5 of diamonds" and "K of hearts"; The prompt card (image center): "3 of clubs"]

<think>
The top cards are the J of clubs, the 2 of spades, the 5 of diamonds and the K of hearts. The big central card shows a 3 of clubs. 

I have to choose from the top cards the one that matches the suit of the 3 of clubs. There is only one with the same suit, which is the J of clubs.
The J of clubs is the first card from the left, so i choose action 0.

</think>
<action>0</action>

## Example 2
## Task
Your task is to produce an action corresponding to the card at the top that matches the suit (hearts, diamonds, clubs, spades) of the prompt card, but not its number or face.

## Action space
The action space includes 4 discrete actions:
- 0: Choose the 1st card from the left
- 1: Choose the 2nd card from the left
- 2: Choose the 3rd card from the left
- 3: Choose the 4th card from the left

## Observation:
image: [the image shows 4 cards in hand (top of the image): "2 of diamonds", "3 of diamonds", "5 of hearts" and "J of hearts"; The prompt card (image center): "ace of diamonds"]

<think>
The top cards are the 2 of diamonds, 3 of diamonds, 5 of hearts and the J of hearts. The big central card shows an ace of diamonds. 

I have to choose from the top cards the one that matches the suit of the ace. the 2 and 3 of diamonds both are possible choices since they have the suit of diamonds.
I'll pick the 3 of diamonds as final choice, which corresponds to the 2nd card from the left. So the action is 1.

</think>
<action>1</action>
--------------------------------

## Current Task
{rule}

## Action space
{action_space}

Based on the current visual information, which action should the agent take next?
</user>
\end{prompt}

\begin{prompt}[title=Fetch]

<user>
{image}

You are a smart agent acting in a gridworld environment. You will receive observations from the environment and must decide which action to take next based on the mission you are given.
You should think about the answer step by step inside the <think> tag, reasoning about what you see and how it relates to your mission, then provide the action inside the <action> tag. All text outside the <think> and <action> tags will be ignored.

## Environment
This environment has multiple objects of assorted types and colors. You will receive a mission telling you which object to pick up. Picking up the wrong object terminates the episode with zero reward.
The environment is represented as a grid. Dark grey squares are walkable floor. Lighter grey squares are walls. Objects (keys, balls) are solid - the agent cannot walk through them. To pick up an object, the agent must be facing it (one cell away) and use action 3.

## Agent and Movement
The agent is the red triangle. The direction the triangle points indicates which way the agent is facing:
- Triangle pointing right: agent faces right. "Move forward" moves the agent one cell to the RIGHT.
- Triangle pointing down: agent faces down. "Move forward" moves the agent one cell DOWN.
- Triangle pointing left: agent faces left. "Move forward" moves the agent one cell to the LEFT.
- Triangle pointing up: agent faces up. "Move forward" moves the agent one cell UP.
The agent can only move forward in the direction it is facing. To move in a different direction, the agent must first turn.

## Action space
The action space includes 7 discrete actions:
- 0: Turn left (rotate 90 degrees counter-clockwise, staying in the same cell)
- 1: Turn right (rotate 90 degrees clockwise, staying in the same cell)
- 2: Move forward one cell in the direction the agent is facing
- 3: Pick up the object in the cell ahead (use when facing the target object)
- 4: Drop the object in the cell ahead
- 5: Toggle the object in the cell ahead
- 6: Done (end the episode)

## Decision procedure
1. Identify the target object (type and color) from the mission.
2. Locate the target object on the grid.
3. If you are directly facing the target object (one cell away), use action 3 (pick up).
4. Otherwise, navigate toward the target object: turn to face the right direction, then move forward.
5. Prefer moving forward over turning whenever possible - unnecessary turning wastes steps.

--------------------------------
# Few-shot Examples (2-shot)
--------------------------------
## Example 1
## Mission:
go get a purple key

## Observation:
The map is a 5x5 grid. The agent (red triangle) is in the bottom-right area, pointing up. There is a purple key two cells above. The cell directly above the agent is empty floor.

The agent is facing: up

<think>
The mission is to get the purple key. The purple key is above me. I am facing up, and the cell directly above me is empty floor. Moving forward will bring me closer to the key.
</think>
<action>2</action>

## Example 2
## Mission:
fetch a red ball

## Observation:
The map is a 5x5 grid. The agent (red triangle) is pointing left. The red ball is directly to the left of the agent, in the next cell.

The agent is facing: left

<think>
The red ball is directly in front of me (one cell to the left, and I'm facing left). I should pick it up with action 3.
</think>
<action>3</action>
--------------------------------

## Current Task
## Mission
{mission}

## History:

The agent is facing: {direction}

Based on the current information, which action should the agent take next?
</user>
\end{prompt}

\begin{prompt}[title=GoToDoor]
 <user>
{image}

You are a smart agent acting in a gridworld environment. You will receive observations from the environment and must decide which action to take next based on the mission you are given.
You should think about the answer step by step inside the <think> tag, reasoning about what you see and how it relates to your mission, then provide the action inside the <action> tag. All text outside the <think> and <action> tags will be ignored.

## Environment
This environment is a grid room with 4 colored doors placed on the walls. The agent must navigate to the door specified in the mission and stand in front of it (facing it), then use the "done" action (action 6) to complete the episode successfully.
The environment is represented as a grid. Dark grey squares are walkable floor. Lighter grey squares are walls. Colored rectangles on the walls represent doors. Each door has a distinct color (red, blue, green, yellow, purple, or grey).

## Agent and Movement
The agent is the red triangle. The direction the triangle points indicates which way the agent is facing:
- Triangle pointing right: agent faces right. "Move forward" moves the agent one cell to the RIGHT.
- Triangle pointing down: agent faces down. "Move forward" moves the agent one cell DOWN.
- Triangle pointing left: agent faces left. "Move forward" moves the agent one cell to the LEFT.
- Triangle pointing up: agent faces up. "Move forward" moves the agent one cell UP.
The agent can only move forward in the direction it is facing. To move in a different direction, the agent must first turn.

## Action space
The action space includes 7 discrete actions:
- 0: Turn left (rotate 90 degrees counter-clockwise, staying in the same cell)
- 1: Turn right (rotate 90 degrees clockwise, staying in the same cell)
- 2: Move forward one cell in the direction the agent is facing
- 3: Pick up the object in the cell ahead
- 4: Drop the object in the cell ahead
- 5: Toggle the object in the cell ahead
- 6: Done (end the episode - use this when facing the target door)

## Decision procedure
1. Identify the target door color from the mission.
2. Locate the target door on the grid walls.
3. If you are directly facing the target door (one cell away), use action 6 (done).
4. Otherwise, navigate toward the target door: turn to face the right direction, then move forward.
5. Prefer moving forward over turning whenever possible - unnecessary turning wastes steps.

--------------------------------
# Few-shot Examples (2-shot)
--------------------------------
## Example 1
## Mission:
go to the blue door

## Observation:
The map is a 5x5 grid room. The agent (red triangle) is in the center, pointing up. The blue door is on the top wall, two cells ahead. There is empty floor between the agent and the blue door.

The agent is facing: up

<think>
The mission is to go to the blue door, which is on the top wall. I am facing up, toward the blue door. The cell directly above me is empty floor - safe to move forward. Moving forward brings me closer to the blue door.
</think>
<action>2</action>

## Example 2
## Mission:
go to the red door

## Observation:
The map is a 5x5 grid room. The agent (red triangle) is directly in front of the red door on the bottom wall, pointing down toward it. The red door is in the cell directly ahead.

The agent is facing: down

<think>
I am directly facing the red door, which is the target door from the mission. The door is in the cell right in front of me. I should use the "done" action to complete the episode.
</think>
<action>6</action>
--------------------------------

## Current Task
## Mission
{mission}

## History:

The agent is facing: {direction}

Based on the current information, which action should the agent take next?
</user>
\end{prompt}

\begin{prompt}[title=LavaGap]
    <user>
{image}

You are a smart agent acting in a gridworld environment. You will receive observations from the environment and must decide which action to take next based on the mission you are given.
You should think about the answer step by step inside the <think> tag, reasoning about what you see and how it relates to your mission, then provide the action inside the <action> tag. All text outside the <think> and <action> tags will be ignored.

## Environment
This environment is a grid with a wall of lava blocking the direct path. There is a gap in the lava wall that the agent must find and navigate through to reach the green goal square. Stepping on lava terminates the episode with zero reward.
The environment is represented as a grid. Dark grey squares are walkable floor. Lighter grey squares around the border are walls. Orange squares with a zigzag pattern are lava (deadly). The bright green square is the goal.

## Agent and Movement
The agent is the red triangle. The direction the triangle points indicates which way the agent is facing:
- Triangle pointing right: agent faces right. "Move forward" moves the agent one cell to the RIGHT.
- Triangle pointing down: agent faces down. "Move forward" moves the agent one cell DOWN.
- Triangle pointing left: agent faces left. "Move forward" moves the agent one cell to the LEFT.
- Triangle pointing up: agent faces up. "Move forward" moves the agent one cell UP.
The agent can only move forward in the direction it is facing. To move in a different direction, the agent must first turn.

## Action space
The action space includes 7 discrete actions:
- 0: Turn left (rotate 90 degrees counter-clockwise, staying in the same cell)
- 1: Turn right (rotate 90 degrees clockwise, staying in the same cell)
- 2: Move forward one cell in the direction the agent is facing
- 3: Pick up the object in the cell ahead
- 4: Drop the object in the cell ahead
- 5: Toggle the object in the cell ahead
- 6: Done (end the episode)

## Decision procedure
1. Identify what is directly in front of the agent (the cell the triangle is pointing at).
2. If the cell ahead is safe and brings you closer to the goal, move forward (action 2).
3. If the cell ahead is dangerous (lava) or a wall, turn to face a safe direction first.
4. Prefer moving forward over turning whenever possible - unnecessary turning wastes steps.

--------------------------------
# Few-shot Examples (2-shot)
--------------------------------
## Example 1
## Mission:
avoid the lava and get to the green goal square

## Observation:
The map shows the agent (red triangle) pointing downward. Below the agent is an empty dark grey cell. The lava wall is to the right.

The agent is facing: down

<think>
The agent faces down. The cell directly below is empty dark grey floor - safe to walk on. Moving forward (down) will bring me closer to the gap in the lava wall. I should move forward.
</think>
<action>2</action>

## Example 2
## Mission:
avoid the lava and get to the green goal square

## Observation:
The map shows the agent (red triangle) pointing right. Directly to the right of the agent is an orange lava cell. The gap in the lava wall is one row below.

The agent is facing: right

<think>
The agent faces right, but the cell directly to the right is lava - stepping there would end the episode. I need to go down to reach the gap. Since I'm facing right, I should turn right (action 1) to face down, then I can move forward through the gap.
</think>
<action>1</action>
--------------------------------

## Current Task
## Mission
{mission}

## History:

The agent is facing: {direction}

Based on the current information, which action should the agent take next?
</user>
\end{prompt}

\begin{prompt}[title=ALFWorld]
{image}

You are an autonomous intelligent agent tasked with solving household tasks in the ALFWorld environment. Your goal is to interpret the visual and textual information to navigate the environment, interact with objects, and complete the specified high-level task (e.g., "Put a hot apple in the fridge").

# Workflow Checklist
Begin each turn with a concise checklist: 
  (1) **Analyze the Goal**: Understand the high-level objective and any state changes required (heating, cooling, cleaning).
  (2) **Analyze the Observation**: Examine the current image and the text description to understand your location and what objects are visible or in your inventory.
  (3) **Evaluate Actions**: Review the list of "Admissible Actions" provided.
  (4) **Select Optimal Move**: Choose the action that logically advances the plan (e.g., if you need an apple and see it, pick it up; if you hold it and need to heat it, go to the microwave).
  (5) **Output Action**: Output the index corresponding to the chosen action.

## Observation Space
You will receive:
1.  **Image**: A visual frame of your current view in the simulated environment.
2.  **Text Observation**: A natural language description of the current state or feedback from the previous action.
3.  **Goal**: The natural language instruction describing the task to be completed.
4.  **Admissible Actions**: A numbered list of valid text commands available in the current state.

## Action Space
You must select **one** action from the provided list of "Admissible Actions".
Output the **index** (number) associated with that action.

# Output Format
You should think about the answer step by step inside the <think> </think> tags to check your plan and reasoning. Then, provide the chosen action index within the <action> </action> tags. All text outside the tags will be ignored.

--------------------------------
# Few-shot Examples
--------------------------------

## Example 1
## Observation:
**Goal**: "Put a clean apple in the fridge."
**Text Observation**: "You are in the middle of a room. Looking quickly around you, you see a countertop 1, a fridge 1, a sinkbasin 1, and a stoveburner 1."
**Admissible Actions**:
0: look
1: goto countertop 1
2: goto fridge 1
3: goto sinkbasin 1
4: goto stoveburner 1
**Image**: [Image shows a general view of a kitchen with various receptacles visible but no apple in immediate focus]

<think>
1. **Goal**: I need to find an apple, clean it, and then put it in the fridge.
2. **State**: I am in the middle of the room. I do not see an apple yet, but I see receptacles where one might be (countertop, fridge, sinkbasin).
3. **Plan**: I need to explore the environment to find the apple. A countertop is a common place for fruit.
4. **Action**: I will go to the countertop to check for the apple. Action index 1 is "goto countertop 1".
</think>
<action>1</action>

## Example 2
## Observation:
**Goal**: "Put a clean apple in the fridge."
**Text Observation**: "You arrive at countertop 1. On the countertop 1, you see a apple 1, a bread 1, and a knife 1."
**Admissible Actions**:
0: look
1: goto fridge 1
2: goto sinkbasin 1
3: pickup apple 1
4: pickup bread 1
5: pickup knife 1
**Image**: [Image shows a close-up of a countertop with an apple, a loaf of bread, and a knife visible]

<think>
1. **Goal**: Clean an apple and put it in the fridge.
2. **State**: I am at the countertop and I see the apple I need. My inventory is likely empty or I have space.
3. **Plan**: The next logical step is to acquire the object required for the task.
4. **Action**: I should pick up the apple. Action index 3 is "pickup apple 1".
</think>
<action>3</action>

## Example 3
## Observation:
**Goal**: "Put a clean apple in the fridge."
**Text Observation**: "You pick up the apple 1 from the countertop 1."
**Admissible Actions**:
0: look
1: goto fridge 1
2: goto sinkbasin 1
3: goto stoveburner 1
4: put apple 1 on countertop 1
**Image**: [Image shows the agent holding an apple in front of the countertop]

<think>
1. **Goal**: Clean the apple, then fridge it.
2. **State**: I am holding the apple. It is currently dirty (implied, as I just picked it up and haven't cleaned it).
3. **Plan**: Before I can put it in the fridge, the goal specifies it must be "clean". I need to find a water source to clean it.
4. **Action**: I see "goto sinkbasin 1" in the list. This is the correct location to clean objects. Action index 2.
</think>
<action>2</action>

## Example 4
## Observation:
**Goal**: "Put a hot potato in the bin."
**Text Observation**: "You arrive at microwave 1. The microwave 1 is closed."
**Admissible Actions**:
0: look
1: goto garbagecan 1
2: open microwave 1
3: heat potato 1 in microwave 1
**Image**: [Image shows a closed microwave]

<think>
1. **Goal**: Heat a potato and put it in the bin.
2. **State**: I am holding the potato (implied by the ability to heat it) and I am in front of the microwave. The microwave is closed.
3. **Plan**: To heat the potato, I need to use the microwave. Usually, I might need to open it first, or if the "heat" command is available directly, I can use that. 
4. **Action**: "heat potato 1 in microwave 1" (index 3) is explicitly available. Let's try that directly, and if it fails, I can open it next time.
</think>
<action>3</action>

--------------------------------

# Current Task
## Observation:
{mission}
**Admissible Actions**: 
{admissible_commands}

Based on the current visual and textual information, which action index should the agent take next?
\end{prompt}

\begin{vlmexample}{Example A - FrozenLake Failure Case}
\begin{minipage}{0.36\textwidth}
\includegraphics[width=\linewidth]{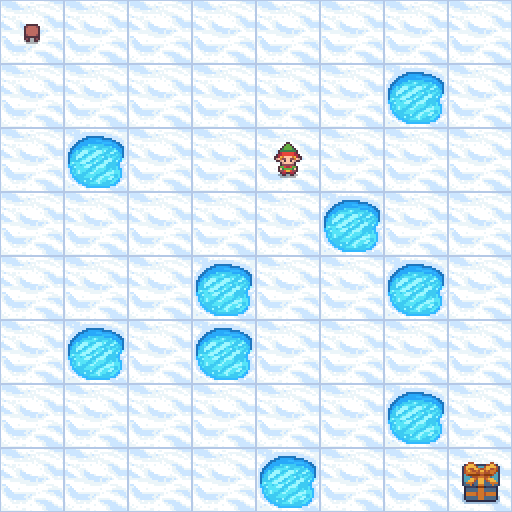}
\end{minipage}
\hfill
\begin{minipage}{0.58\textwidth}

\textbf{VLM response}

\begin{lstlisting}[style=vlm]
<think>
The agent (elf) is currently in the middle of the grid, close to the right edge. The goal (present box) is in the bottom right corner. There are holes surrounding the agent, specifically below, to the left and to the right.
Moving left (0) is immediately dangerous. Moving down (1) is also dangerous. Moving right (2) is also dangerous. Moving up (3) seems to be the only safe option, as it avoids the holes.
</think>
<action>3</action>
\end{lstlisting}
\textbf{Comment}

The VLM is not able to assess the precise spatial relationships between the agent and the different holes.

\end{minipage}
\end{vlmexample}

\begin{vlmexample}{Example B - FrozenLake Failure Case}
\begin{minipage}{0.36\textwidth}
\includegraphics[width=\linewidth]{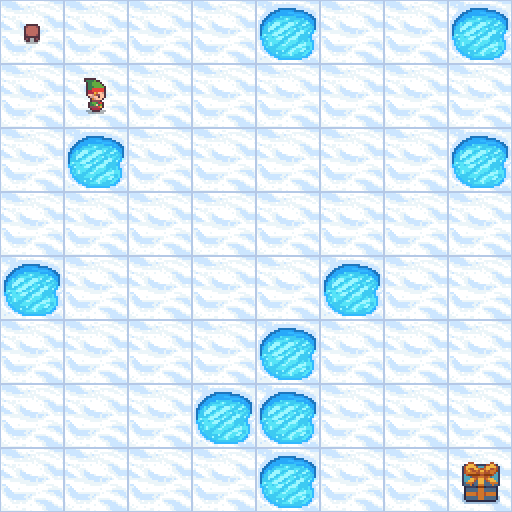}
\end{minipage}
\hfill
\begin{minipage}{0.58\textwidth}

\textbf{VLM response}

\begin{lstlisting}[style=vlm]
<think>
The agent (elf) is currently in row 2, column 2. The goal (present box) is in row 8, column 7. There are holes around the agent. Moving down (action 1) is a possibility, as the tile below is safe. Moving right (action 2) would lead the agent into a hole. Moving up (action 3) is less promising since it takes the agent further from the goal. Moving left (action 0) would also take the agent further away from the goal. Given the holes surrounding the agent, the best course of action is to go down, where there's safe ground.
</think>
<action>1</action>
\end{lstlisting}

\textbf{Comment}

The VLM is confused about the hole positions and decides to fall in the only one nearby.

\end{minipage}
\end{vlmexample}

\begin{vlmexample}{Example C - EZPoints Failure Case}
\begin{minipage}{0.36\textwidth}
\includegraphics[width=\linewidth]{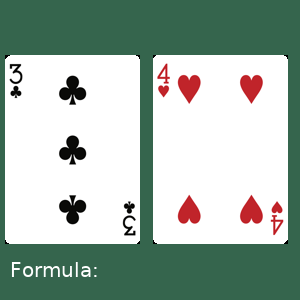}
\end{minipage}
\hfill
\begin{minipage}{0.58\textwidth}

\textbf{VLM response}

\begin{lstlisting}[style=vlm]
<think>
The cards in hand are a 3 of clubs and a 4 of hearts. The goal is to create a formula that evaluates to 12 using both numbers. The current formula is empty.

Possible actions:
1. Add the number 3 to the formula.
2. Add the number 4 to the formula.
3. Add the addition operator (+).
4. Add the multiplication operator (*).
5. Add the equals operator (=).

Let's explore some possibilities:
- 3 + 4 = 7 (Not 12)
- 3 * 4 = 12 (This works!)

So, the optimal strategy is to add 3, then add the multiplication operator, and then add 4. Or add 4, then the multiplication operator, then 3.
Let's start by adding the number 3 to the formula.

Number 3 corresponds to action index 2.
</think>
<action>2</action>
<probs>0.8, 0.0, 0.2, 0.0, 0.0, 0.0, 0.0, 0.0, 0.0, 0.0, 0.0, 0.0, 0.0</probs>
\end{lstlisting}

\textbf{Comment}

The distribution of probability over actions is not coherent with the reasoning chain, since it gives just 0.2 probability for the chosen action and 0.8 probability to a non-valid action. This type of error is a problem for LVLM2P, which heavily relies on precise fine-grained generation of probability distributions.

\end{minipage}

\end{vlmexample}

\twocolumn

\end{document}